\documentclass[final,5p,times,twocolumn]{elsarticle}

\usepackage{times}
\usepackage{epsfig}
\usepackage{graphicx}
\usepackage{amsmath}
\usepackage{amssymb}
\usepackage{makecell}
\usepackage{tabularx}
\usepackage{color}
\journal{Neural Networks}

\begin{document}

\begin{frontmatter}

\title{SIRe-Networks: Convolutional Neural Networks Architectural Extension for Information Preservation via Skip/Residual Connections and Interlaced Auto-Encoders}

\author[inst1]{Danilo Avola}
\author[inst1]{Luigi Cinque}
\author[inst1]{Alessio Fagioli}
\author[inst2]{Gian Luca Foresti}

\affiliation[inst1]{organization={Department of Computer Science, Sapienza University},%Department and Organization
	addressline={Via Salaria 113}, 
	city={Rome},
	postcode={00138}, 
	country={Italy}
}

\affiliation[inst2]{organization={Department of Mathematics, Computer Science and Physics, University of Udine},%Department and Organization
	addressline={Via delle Scienze 20}, 
	city={Udine},
	postcode={33100}, 
	country={Italy}
}

\begin{abstract}
Improving existing neural network architectures can involve several design choices such as manipulating the loss functions, employing a diverse learning strategy, exploiting gradient evolution at training time, optimizing the network hyper-parameters, or increasing the architecture depth. The latter approach is a straightforward solution, since it directly enhances the representation capabilities of a network; however, the increased depth generally incurs in the well-known vanishing gradient problem. In this paper, borrowing from different methods addressing this issue, we introduce an interlaced multi-task learning strategy, defined SIRe, to reduce the vanishing gradient in relation to the object classification task. The presented methodology directly improves a convolutional neural network (CNN) by preserving information from the input image through interlaced auto-encoders (AEs), and further refines the base network architecture by means of skip and residual connections. To validate the presented methodology, a simple CNN and various implementations of famous networks are extended via the SIRe strategy and extensively tested on five collections, i.e., MNIST, Fashion-MNIST, CIFAR-10, CIFAR-100, and Caltech-256; where the SIRe-extended architectures achieve significantly increased performances across all models and datasets, thus confirming the presented approach effectiveness.
\end{abstract}

\begin{keyword}
neural network architectures \sep multi-task learning \sep deep learning \sep object classification
\end{keyword}

\end{frontmatter}

\section{Introduction}
Neural networks and, in particular, deep architectures, perform well on heterogeneous and complex tasks such as medical image analysis \cite{gridach2021pydinet,oksuz2020deep,avola2021multimodal}, person re-identification \cite{wu2020end,tang2020person,avola2020bodyprint} or emotion recognition \cite{zhao2021combining,liang2019unsupervised,avola2020deep}. 
A common objective that has a particularly vast application breadth and spans across many of these fields is the image classification. In fact, nowadays, many datasets focusing either on specific categories, e.g., flowers \cite{nilsback2008automated} or cars \cite{krause20133d}, or comprising completely different object categories \cite{krizhevsky2009learning,deng2009imagenet}, are available. Among this plethora of collections, a simple subdivision can be done on the number of classes contained in a dataset, resulting in increasingly more complex collections. Although the latter can be preferred to fully highlight a method's effectiveness, easier collections can still be useful. For instance, datasets with a low number of categories such as MNIST \cite{lecun1998mnist} and Fashion-MNIST \cite{xiao2017fashion}, on which even simple networks can obtain significant classification results in a short time \cite{keysers2007comparison}, are generally used to validate new ideas such as multi-resolution knowledge distillation to detect anomalies \cite{salehi2021multiresolution} or sparse model-agnostic meta-learning (MAML) to achieve continual learning on a single model \cite{von2021learning}. Datasets containing a moderate amount of diverse categories such as CIFAR-100 \cite{krizhevsky2009learning} and Caltech-101 \cite{fei2006one}, on the other hand, start to show the limitations even of renown and effective models such as ResNet \cite{he2016deep}, that can reach, for example, an error rate of $\approx$22\% on the former collection. 
Nevertheless, there are already solutions that improve such models. For instance, the scheme in \cite{ren2019feature} modifies the internal connections of residual blocks to increase their reuse of the input features, reducing the ResNet error rate by $\approx$2\%. In  \cite{foret2021sharpness}, instead, the authors entirely change the training procedure via a sharpness-aware minimization (SAM) approach, which allows them to implement an efficient gradient descent training procedure by simultaneously minimizing loss value and loss sharpness, and results in a sensibly lower error rate of 7.44\% on the image classification task. 
A similar behavior can also be noticed for the increasingly more complex datasets, such as Caltech-256 \cite{griffin2007caltech} and ImageNet \cite{deng2009imagenet}, that can reach up to 1000 different classes. In fact, on ImageNet, a ResNet achieves $\approx$21\% error rate, while more recent models can reduce this rate close or down to the single digit. For example, the method presented in \cite{shinozaki2021biologically}, which obtains an error rate of 12.28\% on ImageNet, substitutes the backpropagation by an unsupervised approach based on a competitive "winner takes all`` mechanism, implemented as activation function to sparsely update the internal weights of the model. Specifically, this approach leverages exclusively feed forward information to emulate biologically plausible deep neural networks. The work introduced in \cite{dai2021coatnet}, instead, achieves a 9.12\% error rate by skillfully crafting a method that merges convolutional networks and transformers. In particular, they show how a simple relative attention approach can naturally unify both depthwise convolutions and self-attention. Then, by vertically stacking this newfound operation, they can drastically improve performance even on a complex collection such as ImageNet.

A separate but non-negligible aspect of datasets containing a medium or a high number of classes is the time required to train a network. Having more classes, samples, or images with increased width and height (e.g., going from a 28$\times$28 to a 224$\times$224 shape when using the MNIST and ImageNet datasets, respectively) generally requires the design of more complex architectures to achieve higher performance, which naturally translates in a given model requiring a considerably longer training time before it reaches convergence on such a collection.
Regardless, solutions to address these aspects either reduce the number of images simultaneously fed to the model, i.e., the batch size, at the expense of a longer time to complete a training epoch, or use higher-performing hardware with conspicuous amounts of RAM and processing speed, such as tensor processing units (TPUs), to limit eventual training time increases. As a consequence, many approaches are constantly being developed even on the more complex datasets such as ImageNet \cite{zhao2021learnable,cheng2020parametric,wei2020visual}. 

On a different note, while an ever increasing performance is generally obtained by newly developed systems, improving existing architectures can be a daunting experience due, mostly, to the high number of aspects to be accounted for, when designing a model.
Among the many available options, some of the ideas that are being actively explored concern the use of different loss functions, regularization and normalization techniques as well as architectural innovations through, for example, multi-path information processing \cite{khan2020survey}. 
Focusing on methodologies exploiting different loss functions, an interesting and effective solution is associated to the use of composite functions \cite{zhang2019learning,kim2020proxy} to represent, for instance, different outputs to be optimized by a given network, as usually happens in a multi-task learning scenario \cite{chen2020multi,tang2019pose}. 
Such approaches can be further improved by applying weights to the various loss components, allowing a network to focus more on a specific task among those that are being optimized \cite{park2019continual}.
Moreover, further architecture refinements might involve employing intermediate loss functions to alter the gradient evolution; 
a strategy that enables a model to retain better and more meaningful features along the various architecture layers, and allows to obtain improved results on the addressed task \cite{szegedy2015going}.
A different strategy to improve the performance of a network, often used in conjunction with a custom loss function, lies instead in regularization \cite{lui2021improving, liu2019spectral} and normalization techniques \cite{zhang2020exemplar}, where internal weights are modified to enhance the abstraction capabilities of a model via direct manipulation. 
Well-known examples of such approaches comprise a dropout strategy \cite{srivastava2014dropout}, to reduce overfitting over the training dataset distribution by omitting random units at training time for more robust input representations \cite{huang2019prior}; 
and batch normalization \cite{binkowski2019batch,cui2020towards}, to facilitate deep networks training by normalizing a given batch according to a specific strategy such as the batch mean and variance \cite{ioffe2015batch}. 
While these approaches can alleviate issues such as the vanishing/exploding gradient, or directly improve a network performance, a crucial factor is represented by the internal structure of an architecture. 
As a matter of fact, many recent works \cite{khan2020survey} are focusing on architectural ideas, starting from clever configurations such as the inception layers \cite{szegedy2015going}, or skip \cite{ronneberger2015u} and residual connections \cite{he2016deep}.
Indeed, by defining multiple paths inside a given architecture (e.g., inception layer) it is possible to grasp more (or different) details when analyzing an input \cite{wang2020multi};
whereas using extra connections inside a model allows to strengthen feature propagation and reuse and, consequently, obtain improved performances with respect to simpler networks \cite{huang2017densely}.

Inspired by these solutions, in this paper we focus on the image classification task and present the SIRe methodology, where an interlaced multi-task learning approach is exploited jointly with skip and residual connections to improve a convolutional neural network performances. 
Specifically, following the rationale behind GoogleNet \cite{szegedy2015going}, where intermediate classifications are used to mitigate the vanishing gradient problem, we implement a simple CNN model and extend its architectural design to have intermediate auto-encoders that enable the interlaced multi-task learning. 
More accurately, these intermediate tasks require the network to recreate the input image so that information is preserved throughout the network, ultimately providing improved features to the classification layers.
To further enhance both the intermediate and final tasks, we also borrow residual and skip connections from ResNet \cite{he2016deep} and U-Net architectures \cite{ronneberger2015u}, respectively, so that both the classification accuracy and input image reconstruction can obtain higher results with respect to the base model. 
To evaluate the SIRe methodology, extensive experiments were performed on the MNIST, Fashion-MNIST, CIFAR10, CIFAR100 and Caltech256 datasets by extending both a simple CNN as well as various well-known models through the presented approach. 
As demonstrated by the experimental results, a network is enforced to preserve information from the input image once it is modified via the SIRe methodology; thus resulting in a significant performance improvement on the classification task with respect to the base model, which validates the proposed strategy.

Summarizing, the main contributions of this paper are:
\begin{itemize}
	\item designing an interlaced multi-task learning approach to directly affect the gradient evolution;
	\item exploring some solutions (such as skip and residual connections) that enhance network performances;
	\item extending the presented SIRe methodology to other well-known architectures.
\end{itemize}

\begin{figure*}[t]
	\centering
	\includegraphics[width=\textwidth]{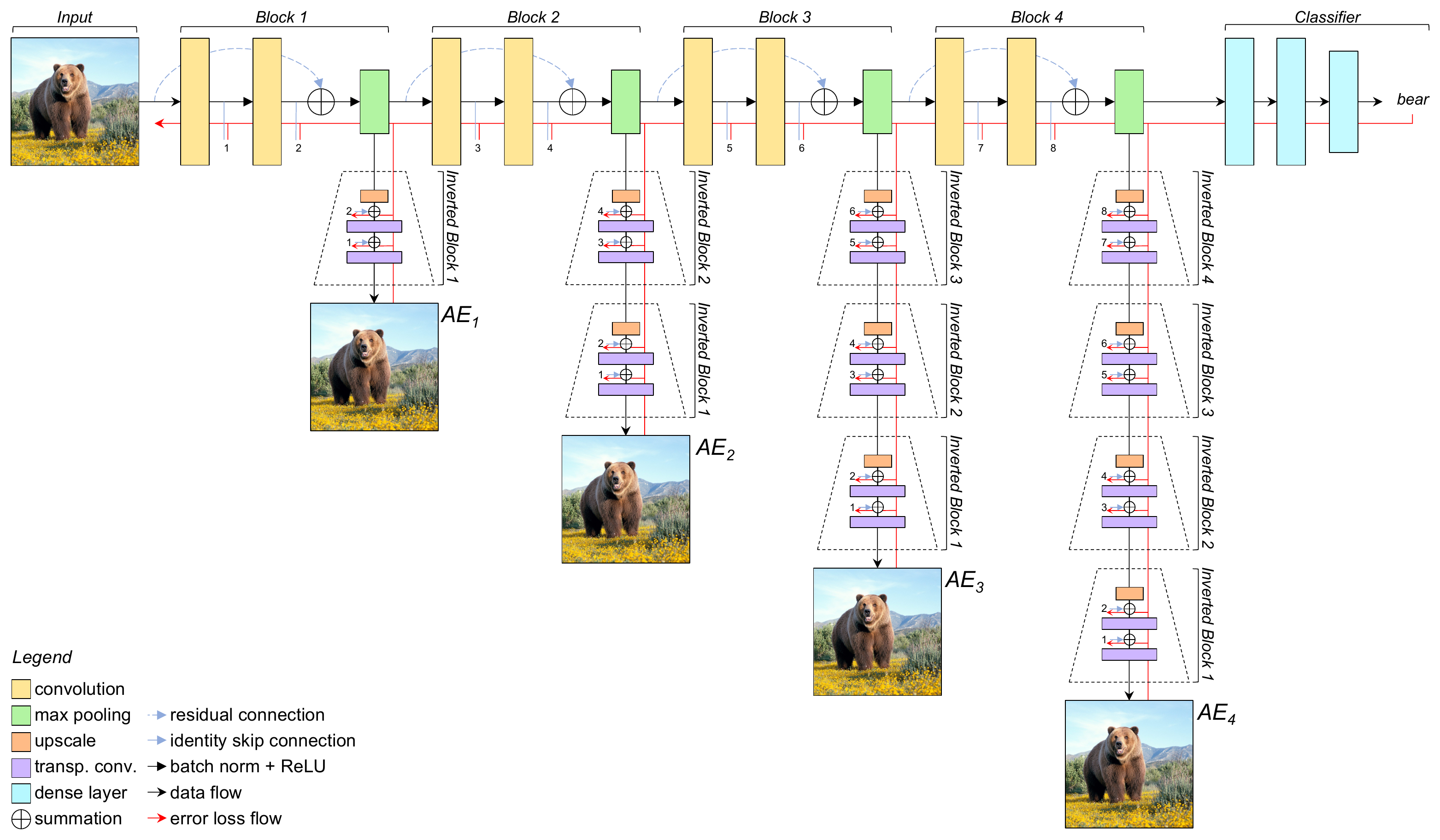}
	\caption{SIRe extended CNN architecture showing the interlaced multi-task learning approach.}
	\label{fig:sire_network}
\end{figure*}

\section{Related Work}
Some of the most effective choices, when designing a new network architecture, entail the loss function to be minimized, which is a direct consequence of the task being addressed; possible regularization and normalization techniques, that might enable a network to converge faster on a given task; and the model architecture itself, where defining wrong operations could result in the network diverging from the required task.

Concerning works exploring loss functions \cite{ouyang2021accelerated,li2019am,xian2020structure,qian2020dr,zheng2019pyramidal,wang2019multi,chen2020contour,choi2020face,szegedy2015going}, an effective strategy is to multiply the loss by a parameter, thus creating a weighted function. 
This approach directly enables the implementation of adaptive loss strategies, where the function is modified according to a specific rationale. 
In \cite{ouyang2021accelerated}, for instance, the relative magnitude between two subsequent losses is exploited to decide when to update the generative adversarial network (GAN) model; while the authors of \cite{li2019am} leverage such parameters to explore possible loss functions and automatically pick the best one. 
Employing parameters can be an effective choice, and other approaches follow a similar reasoning by defining ranking losses to obtain a parametric loss function such as in \cite{xian2020structure} and \cite{qian2020dr}. 
Specifically, the former defines a pair-wise ranking loss based on the input image structure, while the latter designs a distributional ranking loss to better separate positive examples from negative ones. 
A straightforward extension from parametric loss functions is obtained via multi-task loss functions, where different tasks are learned simultaneously. 
In \cite{zheng2019pyramidal} and \cite{wang2019multi}, for example, the cross-entropy and triplet loss functions, in the former, and the self/relative samples similarities, in the latter, are optimized jointly to improve the respective network output. A further refinement for multi-task losses can finally be defined by computing intermediate losses, as described by \cite{chen2020contour} and \cite{szegedy2015going}, where losses for similar tasks are computed along the architecture structure; and from which we took inspiration for the proposed interlaced multi-task learning.

Regularization \cite{yun2019cutmix,zhong2019adversarial,nascimento2019one,ren2019simultaneous} and normalization \cite{luo2019switchable,pan2018two,wu2018group,huang2017centered} techniques can help to achieve a better input abstraction or faster convergence. 
In \cite{yun2019cutmix}, for example, a region dropout is applied to the network input as a data augmentation strategy. 
The authors of \cite{zhong2019adversarial}, instead, develop an elastic regularization strategy to capture differences among diverse inputs; while in \cite{nascimento2019one}, a regularization term is applied to smooth the network output and avoid misclassifying wrong inputs.
Regularization procedures can also be defined for internal layers as shown by \cite{ren2019simultaneous}, where a parametrized regularization is designed to improve the model performance by accounting for both network filters and penalty functions.
Differently from these methods, \cite{pan2018two} and \cite{wu2018group} propose normalization strategies to improve the learning capabilities of an architecture. 
Specifically, \cite{pan2018two} employs instance and batch normalizations to retain, respectively, information invariant to appearance changes (e.g., color) and related to content; while \cite{wu2018group} defines a group normalization, applied on the filters channels, to improve performances even for lower mini-batch sizes. 
Input weights can also be normalized, as shown by \cite{huang2017centered}, where the authors also parametrize such weights to speedup the training procedure. Finally, to directly affect data distributions along their architecture, the authors of \cite{luo2019switchable} present an automatic method to learn the best normalization approach for a given layer. Notice that while many significant advances and interesting approaches have been proposed, in this work we employ standard dropout and batch normalization techniques since they are widely accepted procedures.

This paper also draws inspirations from model architecture variations
\cite{he2016deep,szegedy2015going,ronneberger2015u,zhao2019multi,hao2020labelenc,sun2020classifier,wang2020deep,zhang2020residual,avola2019master} 
\color{black}
that aim to improve the optimization task by analyzing and modifying gradient-derived information. More accurately, the authors of \cite{szegedy2015going} implement intermediate losses to reduce the vanishing gradient problem. 
In \cite{zhao2019multi}, losses computed on internal convolution outputs are exploited to improve the input representation. The authors of \cite{hao2020labelenc} and \cite{sun2020classifier}, instead, show how intermediate losses can reduce the internal representation distance between the components of their respective architectures. Exploiting intermediate loss functions directly affects the gradient evolution during training time; however, more information can also be forwarded inside the architecture through specific connections such as the skip and residual ones. In particular, \cite{he2016deep} and \cite{wang2020deep} apply long skip connections to retain information from previous layers and successfully improve the final output. In \cite{ronneberger2015u} and \cite{zhang2020residual}, instead, residual connections are implemented to forward more information along the architecture, ultimately improving the underlying network abstraction capabilities.

\section{Method}
To present a methodology that consistently improves CNNs performances on the image classification task, we first design a simple network and then extend it through the SIRe technique. In particular, in the proposed approach we implement an interlaced multi-task learning strategy by means of intermediate auto-encoders, to ensure the input image information is preserved; and further improve the model through long skip and short residual connections, to augment the quality of features forwarded throughout the architecture. The presented SIRe-extended CNN architecture is shown in Fig.~\ref{fig:sire_network}.

\subsection{Base CNN Model}
The first step to introduce the SIRe methodology requires the implementation of a simple CNN architecture to be used as a baseline. 
Specifically, starting from the well-known VGG \cite{simonyan2014very} architecture, we define a less powerful model (i.e., using a fewer number of convolutions and dense units) that can be easily extended and used as a baseline to validate the SIRe adaptation. 
Notice that the VGG was selected since it has a straightforward implementation, however any other network can be used and extended through the presented methodology, as also highlighted by the experimental results. 
In more details, the baseline CNN extracts features from an input image through 8 convolutional layers, each followed by a batch normalization, and 4 max pooling layers, placed every 2 convolutions. Subsequently, 3 dense layers are used for the classification task, similarly to the original VGG structure.
Concerning the convolutions, in all layers a kernel size $k=3$ is jointly applied with a padding $p=1$ to ensure the input shape is retained and reduced only via the max pooling layers. Moreover, the number of filters generated by the convolutions is doubled after every max pooling operation, starting from 64 in the first convolution (i.e., 64, 128, 256, and 512). Regarding the dense layers, the first two employ the same number of nodes (i.e., 1024) and a dropout strategy to avoid overfitting over the training dataset; while the third one contains $m$ units, where $|m|$ corresponds to the number of classes to be recognized. 
Finally, the activation function applied to all $l-1$ layers is the rectified linear activation function (ReLU), while the last one uses a softmax function to obtain a probability distribution for the input classification.

\subsection{SIRe Extension}
Core component behind the SIRe adaptation lies in the interlaced multi-task learning strategy, where intermediate auto-encoders are used to extend the architecture and manipulate the gradient. Intuitively, these auto-encoders are tasked with the input image reconstruction as to preserve its attributes inside the convolutions. Notice that by defining such components, the base architecture effectively performs an interlaced multi-task learning, where the original task is the object classification, while the intermediate ones correspond to the input reconstruction. Furthermore, the rationale behind this approach is two-fold. 
First, it can indirectly alleviate the classification task vanishing gradient problem, as was also suggested by the authors of \cite{szegedy2015going}, by defining extra tasks that inject additional gradient error inside the model at training time. Second, it naturally enforces the network to preserve meaningful input characteristics (i.e., features) along the whole architecture; therefore providing an improved input abstraction to the classifier that can then achieve higher performances.
In particular, an auto-encoder is built upon each max pooling layer, i.e., $AE_i$, with $i\in[1,\dots,|max\_pool\_layers|]$, since they correspond to most of the information loss due to the dimensionality reduction. Moreover, in order to allow the correct output generation, every $AE$ follows the underlying architecture structure. Specifically, they invert all max pooling operations and convolution layers, through the upscaling strategy proposed in \cite{ronneberger2015u} and transposed convolutional layers \cite{dumoulin2016guide}, respectively. For example, as shown in Fig.~\ref{fig:sire_network}, the auto-encoder built upon the third max pooling layer will perform 3 upscale operations, to reach the original input image size, interleaved by 6 transposed convolutions (i.e., 2 layers after each upscale) employing the same $k$ and $p$ of the convolutional layers to correctly reinterpret the input.%
%\color{red}
% Notice that a $AE$s inverting the base architecture, built starting from each max pooling operation to preserve most of the information along the network
% Notice that each $AE$ inverting the base architecture is completely distinct from the other ones so that they can preserve a maximal amount of information along the network.
Notice that each $AE$ inverting the base architecture is entirely distinct from the others, so they can all preserve the maximum amount of information within the network structure.
% , i.e., the model illustrated in Fig.~\ref{fig:sire_network} there are four separate $AE$s when implementing the SIRe extension to 
% Notice
% that distinct AEs inverting the base architecture are built start-
% ing from each max pooling operation in order to preserve most
% of the information along the network
\color{black}

Attaching interleaved tasks via auto-encoders helps to inject gradient error during the backpropagation algorithm execution. However, since the intermediate tasks greatly differ from the final output in both structure and purpose, a further enhancement is implemented to improve the information forwarded inside the architecture by means of skip and residual connections. 
In more details, short residual connections are applied before each max pool operation and connect two subsequent convolutions with their input, similarly to \cite{he2016deep}; while long skip connections bond a given convolution with its corresponding transposed convolution in the auto-encoder, to compensate with missing information derived from both the max pool and upscale operations. 
Notice that neither short residual and long skip connections increase the number of parameters or network complexity, since they are implemented through identity maps and simple summations in accordance with the findings of \cite{he2016deep}; therefore resulting in great tools for information propagation and performance improvement. 

\subsection{Loss Function}
A SIRe extended network addressing the classification task, can still be trained end-to-end via classical algorithms such as the stochastic gradient descent (SGD) with backpropagation. 
However, a specific loss function needs to be implemented to fully leverage the presented methodology and correctly merge the intermediate objectives with the original task.
Intuitively, by exploiting the computational graph associated to the network structure, each auto-encoder will focus on specific portions of the architecture and automatically increase the gradient error throughout the model. 
For instance, observing Fig.~\ref{fig:sire_network}, the first auto-encoder will affect only the first two convolutions, while the fourth one will influence the whole architecture during the error backprogation.
What is more, through the backprogation, the interlaced multi-task methodology enforces to retain structural information on the input since all auto-encoders try to correctly reconstruct the original image.
Formally, given a SIRe extended network with several auto-encoders $AE$, the interlaced multi-task loss function is computed as:
\begin{equation}
	\mathcal{L} = -\frac{1}{m}\sum_{i=1}^m y_i\log(\hat{y}_i) + \lambda\sum_{AE}\left(\frac{1}{n}\sum_{j=1}^n (AE_{o_j} - x_j)^2\right),
\end{equation}
where the first term is a cross-entropy loss computed over the $m$ classes, used to classify the input image; while the second member corresponds to a mean squared error loss for each available auto-encoder $AE$, evaluating the input image reconstruction. In particular, $AE_{o_j}$ and $x_j$ indicate, respectively, the $j$-th pixel predicted output of a given auto-encoder $AE$ and input image (i.e., ground truth for the intermediate task); $n$ represents the input image number of pixels; while $\lambda$ is an hyperparameter regulating the interlaced multi-task loss strength over the underlying architecture, empirically set to 0.2 in this work. Notice that due to the diverse nature of the intermediate tasks and final output, a too high value of $\lambda$ might result in the network focusing on the input reconstruction instead of the classification, as demonstrated by the performed experiments. 

\section{Experiments}
Extensive experiments were carried out to fully evaluate the proposed SIRe methodology. 
The datasets and test protocols employed to ensure sound results across all tests, are described in Section \ref{sub:implementation_details}. Quantitative and qualitative ablation studies are discussed in Section \ref{sub:ablation_study}, where in depth details on the described approach are provided. Finally, experiments on SIRe extended networks are presented in Section \ref{sub:sire_networks}, to highlight the methodology effectiveness and its limitations.

\subsection{Implementation Details}
\label{sub:implementation_details}

The SIRe extension was evaluated on five distinct collections, i.e., MNIST \cite{lecun1998mnist}, Fashion-MNIST \cite{xiao2017fashion}, CIFAR-10 \cite{krizhevsky2009learning}, CIFAR-100 \cite{krizhevsky2009learning}, and Caltech-256 \cite{griffin2007caltech}. Firstly, extensive ablation studies of a SIRe-CNN model were performed on CIFAR-100.
This dataset contains 60000 32$\times$32 RGB images spanning over 100 classes. In particular, there are 500 training and 100 test images per class, resulting in the training and test sets containing 50000 and 10000 images, respectively. Moreover, in this work we further split the training set to retain 50 images per class to define a validation set. Therefore, the final collections count 45000, 5000, and 10000 samples, for the training, validation, and test sets. 
These splits were also used to evaluate other SIRe-extended literature networks on a dataset containing a medium amount of categories.
Secondly, MNIST, Fashion-MNIST, and CIFAR-10 datasets were used to validate the SIRe effectiveness on collections with a fewer number of classes. More specifically, MNIST and Fashion-MNIST comprise 70000 28$\times$28 grayscale images of digits from 0 to 9 and common Zalando's fashion articles, respectively. CIFAR-10, instead, consists of 70000 32$\times$32 RGB images of ten diverse object categories. All three datasets were divided into train, validation and test sets containing 50000, 10000, and 10000 samples. Finally, the Caltech-256 dataset was used to assess the SIRe methodology on a collection with a higher number of classes. More specifically, this dataset contains 30607 real-world images of different sizes spanning over 257 classes, i.e., 256 object categories and an additional clutter class. To handle the shape discrepancy, all images are resized to a dimension of 224$\times$224. In this last collection, due to the increased difficulty and lower number of images per class, i.e., a minimum of 80 images, the dataset was divided only into training and test sets using a 70\%/30\% split for a total of 21531 and 9076 samples. 

Regarding the various experimental settings, all models were trained following a protocol similar to \cite{devries2017improved} for tests on CIFAR-10, CIFAR-100, and Caltech-256. 
Specifically, each network was trained for 200 epochs using the SGD algorithm, with an initial learning rate $lr$ set to 0.1, a weight decay of 5e-4, a Nesterov momentum of 0.9, and a batch size ranging from 4 up to 64, depending on the datasets and underlying architecture. Moreover, a scheduler was also implemented to divide the $lr$ by 5 at epochs 60, 120, and 160, so that the gradient update speed would be gradually reduced. 
Contrary to these settings, for MNIST and Fashion-MNIST we trained the models only for 5 epochs, using exclusively the first 10 batches of the training set, i.e., 1280 images. This decision was necessary since even the base models would otherwise reach near perfect classification accuracy after the first training epoch.
In all cases, a simple data augmentation strategy was also applied by means of random horizontal flips. 
Finally, for all models, the weights associated to the epoch with the highest performance on the validation set (or best training epoch in the case of Caltech-256, due to the validation set absence) were used to compute the reported results on the test set. 
Notice that for all experiments, the metrics of choice were the top-1(-5) error percentage, which is computed as $1-\text{rank}$ \#1(\#5) accuracy; and where lower scores correspond to better performances.

Lastly, all networks were implemented through the PyTorch framework and tests were performed using a single GPU, i.e., a GeForce GTX 1070 with 8GB of RAM.

\subsection{Ablation Studies}
\label{sub:ablation_study}

The first experiment explores the sheer effectiveness of each SIRe component, i.e., the skip connections (S), interlaced multi-task learning (I), and residual connections (Re). Performances are summarized in Table~\ref{tab:sire_ablation}. As shown, all configurations can achieve significant top-1 and top-5 classification performances, with the former obtaining substantially higher gains, independently from the SIRe component. 
Notice that this score increment difference is most entirely attributable to classes having similar characteristics (e.g., girl and boy) that can be lost due to the small image format of the dataset, but that would still allow for the correct category to be selected among the first 5 most probable classes.
\begin{table}[t]
	\centering
	\begin{tabular}{|l|c|c|r|c|}
		\hline
		Model & \makecell{Top-1\\Error\%} & \makecell{Top-5\\Error\%} & Params & \makecell{Training\\Time} \\
		\hline
		\hline
		Baseline & 35.44 & 12.11 & 7.9M & $\approx$124' \\
		Re-CNN & 32.21 & 10.78 & 7.9M & $\approx$132' \\
		I-CNN & 29.05  & 9.33 & 14.0M & $\approx$205' \\
		IRe-CNN & 27.65 & 8.97 & 14.0M & $\approx$207' \\
		SI-CNN & 27.83 & 9.01 & 14.0M & $\approx$206' \\
		SIRe-CNN & 26.15 & 8.22 & 14.0M & $\approx$212' \\
		\hline
	\end{tabular}
	\caption{Ablation study on SIRe components. S, I, and Re, indicate skip connections, interleaved multi-task learning, and residual connections, respetively. All models employ the same CNN, which is also reported as Baseline.}
	\label{tab:sire_ablation}
\end{table}
\begin{table}[t]
	\centering
	\begin{tabular}{|l|c|c|c|}
		\hline
		Model & $\lambda$-Value & \makecell{Top-1\\Error\%} & \makecell{Top-5\\Error\%}\\
		\hline
		\hline
		Baseline & - & 32.21 & 10.78 \\
		SIRe-CNN & 0.1 & 31.87 & 10.66 \\
		SIRe-CNN & 0.2 & 26.15 & 8.22 \\
		SIRe-CNN & 0.5 & 99.00 & 99.00 \\
		SIRe-CNN & 1.0 & 99.00 & 99.00 \\
		\hline
	\end{tabular}
	\caption{Ablation study on $\lambda$-value effects. The Baseline model corresponds to the Re-CNN of Table~\ref{tab:sire_ablation}.}
	\label{tab:lambda_ablation}
\end{table}
\begin{figure}[t]
	\centering
	\includegraphics[width=\columnwidth]{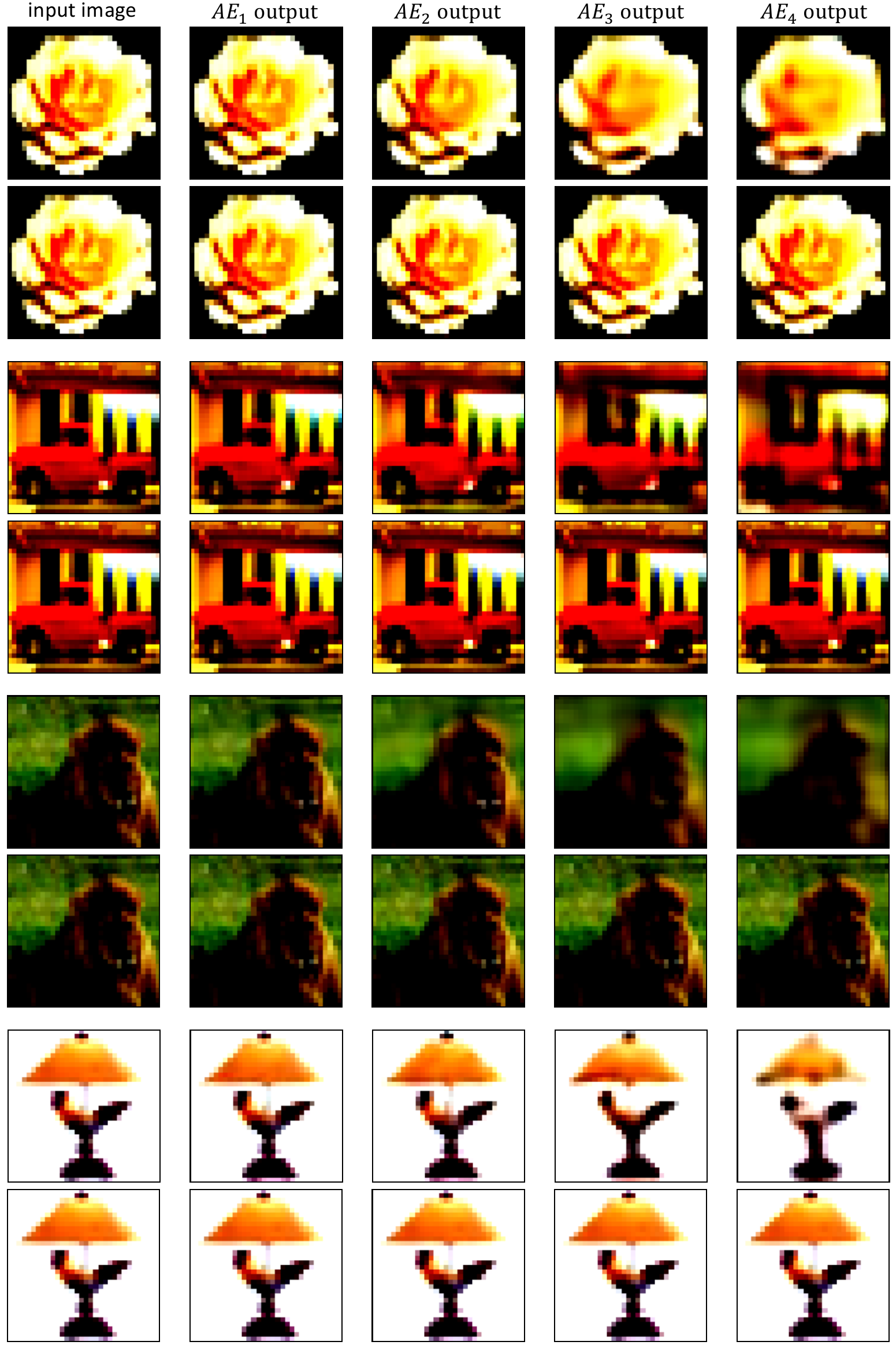}
	\caption{Skip connections effect on the auto-encoders output. The first row for each image corresponds to the IRe-CNN model, while the second row to the full SIRe-CNN. From left to right, the input image and the four auto-encoder outputs.}
	\label{fig:skipconn_study}
\end{figure}
\begin{figure}[t]
	\centering
	\includegraphics[width=\columnwidth]{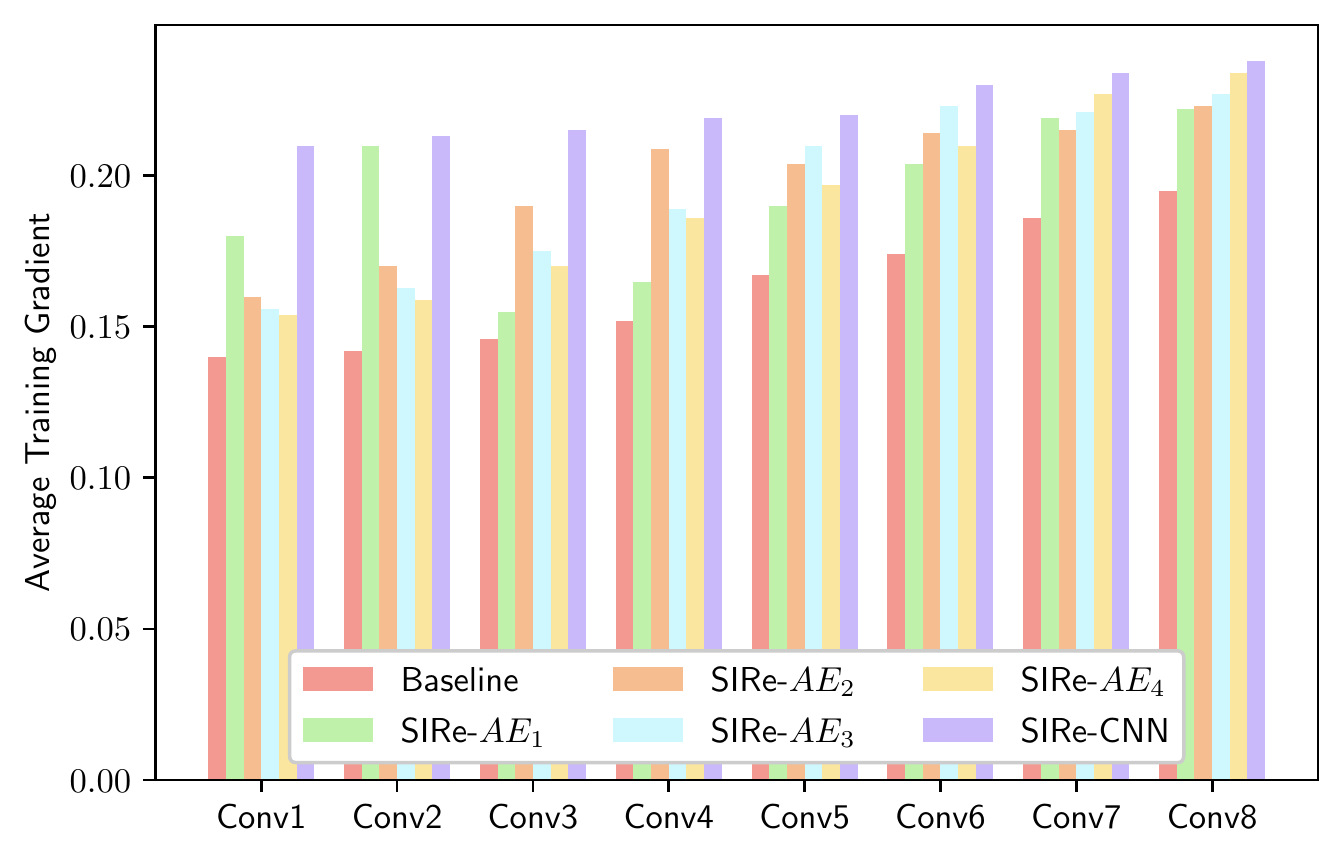}
	\caption{Average gradient inside the 8 convolutional layers for the baseline CNN; auto-encoders $AE_1$, $AE_2$, $AE_3$, $AE_4$ in standalone solutions; and SIRe-CNN. All values are computed during the corresponding model training phase.}
	\label{fig:gradient_backprop}
\end{figure}
\begin{figure*}[t]
	\centering
	\includegraphics[width=\textwidth]{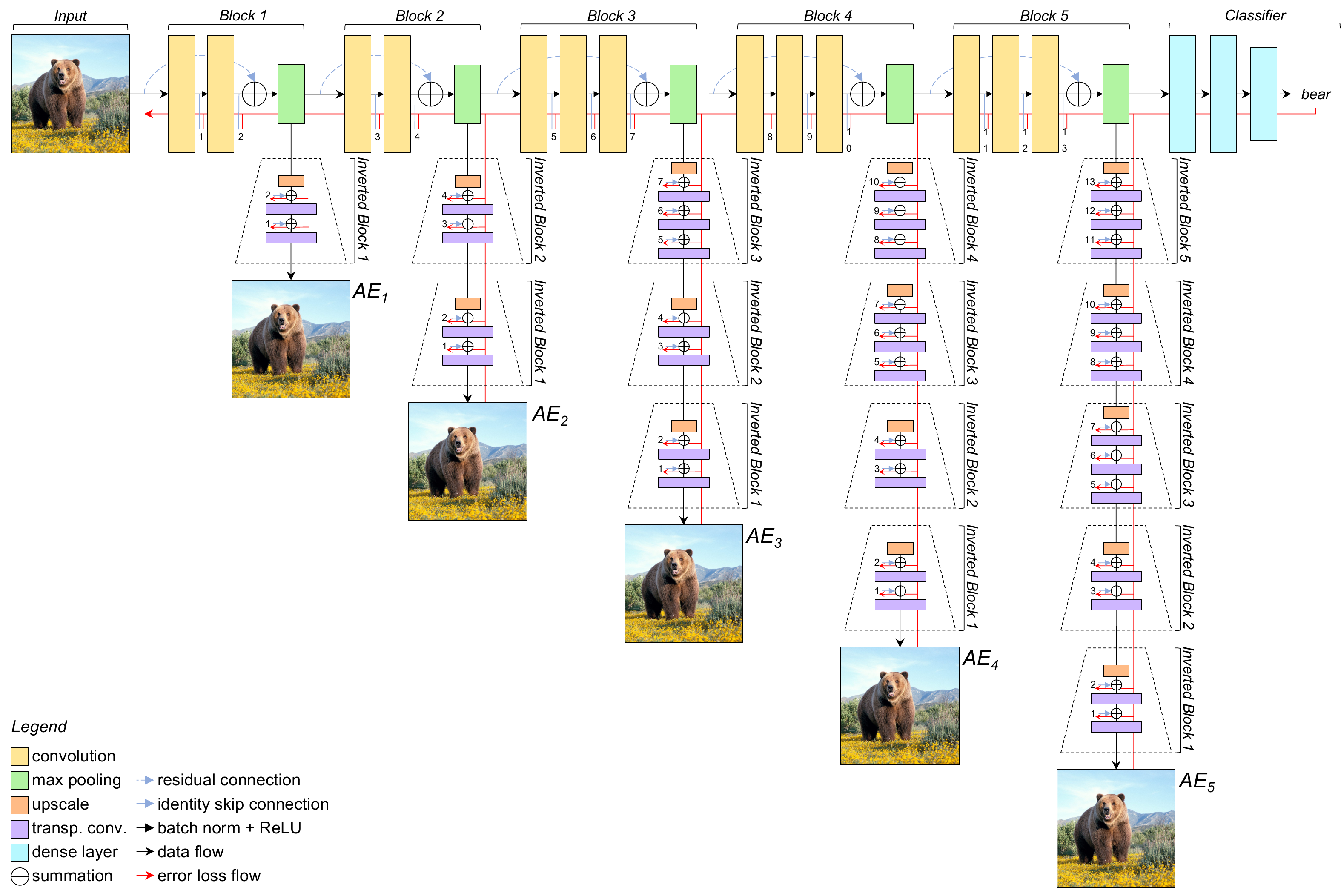}
	\caption{SIRe-VGG16 architecture showing the interlaced multi-task learning approach.}
	\label{fig:sire_vgg16}
\end{figure*}
\begin{figure*}[t]
	\centering
	\includegraphics[width=\textwidth]{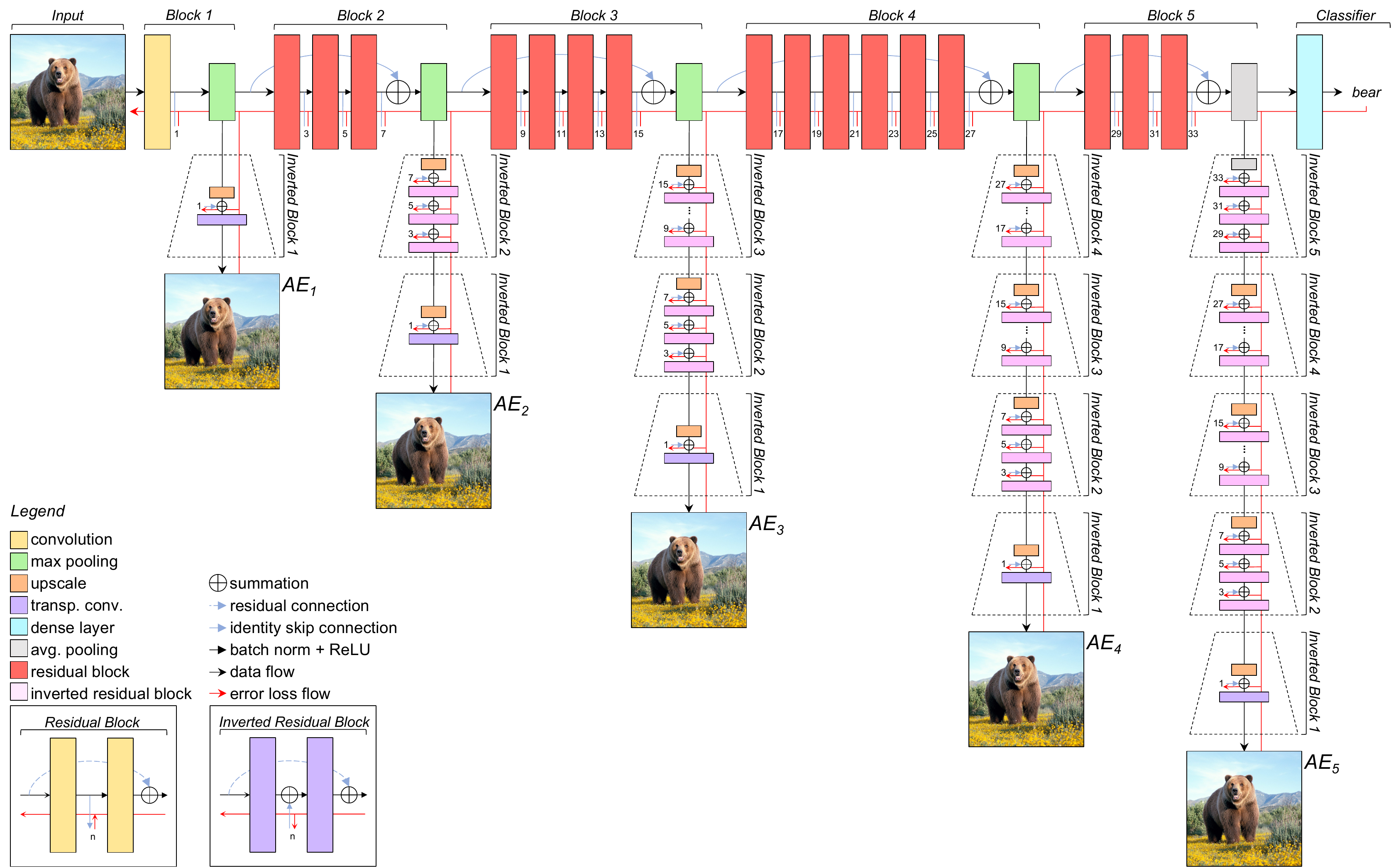}
	\caption{SIRe-ResNet34 architecture showing the interlaced multi-task learning approach.}
	\label{fig:sire_resnet34}
\end{figure*}
What is more, while each extension offers a performance boost, even when applied as a standalone solution, they affect different aspects of the underlying architecture and should, therefore, be used jointly. 
In particular, the Re unit provides a direct upgrade to the base CNN final classification by forwarding residual information along its architecture. The I component offers the highest error reduction among the single extensions, by focusing on the input image structure preservation; however, it also requires more parameters to be implemented and, as a consequence, more time to be trained (i.e., $\approx$ 1.7x more time with respect to the base network). Finally, the S component directly refines the interlaced multi-task outputs by forwarding information to the various $AE$s and, therefore, also reduces the final network error.
This intermediate output refinement can be clearly observed in Fig.~\ref{fig:skipconn_study}. As shown, without implementing skip connections, the max pool and upsample operations result in information loss, especially when moving toward the deeper auto-encoders that, consequently, reconstruct blurred images. Through the S components, instead, all lost input information is retrieved and almost perfect input reconstruction is achieved (i.e., MSE loss $\le$ 3e-6). Nevertheless, while skip connections improve the $AE$s output and enforce the input image structure preservation, they do not guarantee a comprehensive class abstraction. As a matter of fact, the last two images (i.e., lion and lamp) are correctly classified only in the top-5 most probable classes, even though they are still correctly recreated.

The second experiment investigates the $\lambda$-value effects on the training task; where $\lambda=0$ removes the interlaced auto-encoders, while $\lambda\ge 1.0$ takes into account the entire $AE$ error. 
Several significant results for these tests are reported in Table~\ref{tab:lambda_ablation}. As can be seen, using a small $\lambda$-value (i.e., $\le0.2$) enables the underlying CNN to extract more meaningful features and increase its classification performances. Differently, by increasing $\lambda$ over the value 0.2, which was empirically found to be the best amount, the underlying CNN error increases until the classification task does not reach convergence anymore, i.e., the CNN always outputs the same class for any given input; thus resulting in the reported 99\% error. The rationale behind this outcome can be easily explained by the backprogated gradient derived from the interlaced auto-encoders. Specifically, for higher $\lambda$-values, the loss function focuses on the input image reconstruction and all internal parameters are modified to address this task that should, instead, support the classification one.
In practice, the gradient associated to the classification task, due to being affected by the vanishing gradient problem, is completely overridden by the reconstruction error backpropagated by the $AE$s, thus resulting in the reported divergence. 
Nevertheless, by choosing an opportune $\lambda$-value (i.e., $\lambda=0.2$), the right amount of $AE$ error is backpropagated and a performance increase is obtained; indicating that intermediate and diverse tasks (e.g., the proposed image reconstruction) can help to improve the original one.

The third experiment analyzes the amount of information each $AE$ provides to the classification task. An evaluation comprising all possible $AE$ combinations, where each $i$-th $AE_i$ corresponds to those presented in Fig.~\ref{fig:sire_network}, is reported in Table~\ref{tab:ae_ablation}. As shown, by implementing more $AE$s (e.g., $AE_{1,2,4}$) the error decreases sensibly, however more parameters and training time are required as a trade-off for the improved performance. What is more, this compromise is more marked for models employing deeper $AE$s (i.e., $AE_3$ and $AE_4$) since more upscale and transposed convolution operations have to be performed to correctly recreate the input. 
Nevertheless, redundancy by means of shallower $AE$s (i.e., $AE_1$ and $AE_2$) enables for relevant information to be extracted by the base-CNN and allows to obtain better results on the final classification task; thus encouraging the application of multiple $AE$s even though deeper ones have slightly better performance gains. The rationale behind this behavior is also highlighted in Fig.~\ref{fig:gradient_backprop}, where the convolutional layers gradient flow, i.e., the average error propagated in the backward pass of the backpropagation algorithm computed at training time, is reported for each $AE$, as well as the whole SIRe-CNN and base CNN models. As can be seen, while each $AE$ mitigates the vanishing gradient problem (i.e., additional error is propagated along the architecture due to the information preservation encouraged by the reconstruction task) they still do not solve it entirely and, therefore, support the chosen redundancy for a better error propagation.
As a matter of fact, the gradient flow for the SIRe-CNN is more uniform across all convolutions, while it shows error peaks when analyzing the single $AE$s (e.g., SIRe-$AE_1$ has such a peak in convolutional layers 1 and 2).
Notice that applying different $\lambda$-values to each auto-encoder to address the vanishing gradient does not solve the issue due to diminished returns on far layers (e.g., the first convolution is slightly affected by $AE_4$ in comparison with $AE_1$). Moreover, the same issue described in the $\lambda$-value ablation study (i.e., failed convergence) also applies to the single components, forcing the final value to be close to the selected $\lambda=0.2$ to obtain a significant performance boost.
Finally, observe that the interlaced auto-encoders directly affect only the gradient flow of the convolutional layers that are responsible for the extraction of features from an input image. Regardless, gradients associated to the classification layers can also benefit from this manipulation for two reasons. Firstly, they are backpropagated throughout the network together with $\lambda$-regulated (i.e., reduced) errors from the $AE$s. Secondly, more relevant features can be extracted by the network due to the information preserved by the input image reconstruction. Indeed, in accordance with the results shown in Table~\ref{tab:ae_ablation}, performances improve when more information is preserved throughout the architecture, i.e., the error decreases when more $AE$s are implemented. 

\begin{table}
	\centering
	\resizebox{\columnwidth}{!}{
		\begin{tabular}{|l|c|c|r|c|}
			\hline
			Model & \makecell{Top-1\\Error\%} & \makecell{Top-5\\Error\%} & Params & \makecell{Training\\Time} \\
			\hline
			\hline
			Baseline & 35.44 & 12.11 & 7.9M & $\approx$124' \\
			\hline
			SIRe-$AE_1$ & 32.54 & 11.41 & 7.9M & $\approx$125' \\
			SIRe-$AE_2$ & 32.28 & 11.29 & 8.2M & $\approx$126' \\
			SIRe-$AE_3$ & 32.04 & 11.13 & 9.0M & $\approx$133' \\
			SIRe-$AE_4$ & 31.89 & 11.00 & 12.6M & $\approx$195' \\
			\hline
			SIRe-$AE_{1,2}$ & 30.31 & 10.27 & 8.2M & $\approx$127' \\
			SIRe-$AE_{1,3}$ & 30.12 & 10.05 & 9.1M & $\approx$134' \\
			SIRe-$AE_{1,4}$ & 30.07 & 9.71 & 12.6M & $\approx$197' \\
			SIRe-$AE_{2,3}$ & 30.09 & 9.70 & 9.3M & $\approx$138' \\
			SIRe-$AE_{2,4}$ & 29.91 & 9.39 & 12.8M & $\approx$199' \\
			SIRe-$AE_{3,4}$ & 29.55 & 9.31 & 13.7M & $\approx$208' \\
			\hline
			SIRe-$AE_{1,2,3}$ & 26.99 & 8.92 & 9.3M & $\approx$141' \\
			SIRe-$AE_{1,2,4}$ & 26.71 & 8.24 & 12.9M & $\approx$200' \\
			SIRe-$AE_{2,3,4}$ & 26.32 & 8.23 & 14.0M & $\approx$211' \\
			\hline
			SIRe-CNN & 26.15 & 8.22 & 14.0M & $\approx$212' \\
			\hline
		\end{tabular}
	}
	\caption{Ablation study on information propagated by the various auto-encoders defined in the SIRe methodology focused on their depth and their redundancy. All models 
		%\color{red}
		employ $\lambda=0.2$ 
		\color{black}
		for a fair comparison.}
	\label{tab:ae_ablation}
\end{table}

Summarizing, each component of the SIRe methodology can improve the base CNN performance even as a standalone solution. Moreover, provided an opportune $\lambda$-value is selected, implementing multiple intermediate tasks further boosts the original network performance; demonstrating the SIRe extension effectiveness.

\begin{table*}[t]
	\centering
	\begin{tabular}{|l|c|c|c|c|r|c|}
		\hline
		Model & \thead{Batch\\Size} & $\lambda$-value & Top-1 Error\% & Top-5 Error\% & Params & Training Time \\
		\hline
		\hline
		Baseline-CNN & 64 & - & 35.44 & 12.11 & 7.9M & $\approx$119' \\
		SIRe-CNN & 64 & 0.20 & 26.15 & 8.22 & 14.0M & $\approx$202' \\
		\hline
		VGG16 & 64 & - & 29.18 & 13.01 & 138.3M & $\approx$187' \\
		SIRe-VGG16 & 64 & 0.20  & 23.02 & 10.00 & 160.9M & $\approx$336' \\
		\hline
		VGG19 & 64 & - & 28.29 & 11.43 & 143.6M & $\approx$195' \\
		SIRe-VGG19 & 64 & 0.20  & 22.67 & 9.62 & 176.8M & $\approx$370' \\
		\hline
		ResNet34 & 64 & - & 22.17 & 6.29 & 21.8M & $\approx$412' \\
		SIRe-ResNet34 & 64 & 0.21  & 18.35 & 6.01 & 50.4M & $\approx$698' \\
		\hline
		ResNet50 & 64 & - & 21.26 & 5.41 & 25.6M & $\approx$433' \\
		SIRe-ResNet50 & 64 & 0.21  & 17.89 & 5.32 & 59.4M & $\approx$721' \\
		\hline
		GoogleNet & 64 & - & 22.92 & 6.61 & 13.0M & $\approx$317' \\
		SIRe-GoogleNet & 64 & 0.19  & 18.21 & 5.99 & 17.0M & $\approx$380' \\
		\hline
	\end{tabular}
	\caption{Comparison between several literature networks and their SIRe-extended version on the CIFAR100 dataset.}
	\label{tab:sire_networks_cifar100}
\end{table*}

\begin{table}[t]
	\centering
	\resizebox{\columnwidth}{!}{%
		\begin{tabular}{|l|c|c|c|r|c|}
			\hline
			Model & \thead{Batch\\Size} & \thead{Top-1\\Error\%} & \thead{Top-5\\Error\%} & Params & \thead{Training\\Time} \\
			\hline
			\hline
			Baseline-CNN & 64 & 37.28 & 4.50 & 7.9M & $\approx$10'' \\
			SIRe-CNN & 64 & 22.88 & 1.08 & 14.0M & $\approx$18'' \\
			\hline
			VGG16 & 64 & 30.80 & 3.48 & 138.3M & $\approx$13'' \\
			SIRe-VGG16 & 64 & 14.70 & 0.97 & 160.9M & $\approx$23'' \\
			\hline
			VGG19 & 64 & 47.63 & 4.17 & 143.6M & $\approx$15'' \\
			SIRe-VGG19 & 64 & 34.71 & 8.53 & 176.8M & $\approx$24'' \\
			\hline
			ResNet34 & 64 & 14.38 & 0.62 & 21.8M & $\approx$16'' \\
			SIRe-ResNet34 & 64 & 5.22 & 0.27 & 50.4M & $\approx$29'' \\
			\hline
			ResNet50 & 64 & 36.08 & 3.37 & 25.6M & $\approx$23'' \\
			SIRe-ResNet50 & 64 & 18.55 & 2.09 & 59.4M & $\approx$37'' \\
			\hline
			GoogleNet & 64 & 57.77 & 0.99 & 13.0M & $\approx$13'' \\
			SIRe-GoogleNet & 64 & 32.98 & 1.25 & 17.0M & $\approx$16'' \\
			\hline
		\end{tabular}
	}
	\caption{Comparison between several literature networks and their SIRe-extended version on the MNIST dataset.}
	\label{tab:sire_networks_mnist}
\end{table}

\begin{table}[t]
	\centering
	\resizebox{\columnwidth}{!}{%
		\begin{tabular}{|l|c|c|c|r|c|}
			\hline
			Model & \thead{Batch\\Size} & \thead{Top-1\\Error\%} & \thead{Top-5\\Error\%} & Params & \thead{Training\\Time} \\
			\hline
			\hline
			Baseline-CNN & 64 & 24.21 & 0.65 & 7.9M & $\approx$9'' \\
			SIRe-CNN & 64 & 20.04 & 0.63 & 14.0M & $\approx$19'' \\
			\hline
			VGG16 & 64 & 24.53 & 0.75 & 138.3M & $\approx$12'' \\
			SIRe-VGG16 & 64 & 22.32 & 0.75 & 160.9M & $\approx$23'' \\
			\hline
			VGG19 & 64 & 31.44 & 1.11 & 143.6M & $\approx$13'' \\
			SIRe-VGG19 & 64 & 27.11 & 1.29 & 176.8M & $\approx$25'' \\
			\hline
			ResNet34 & 64 & 25.09 & 0.81 & 21.8M & $\approx$18'' \\
			SIRe-ResNet34 & 64 & 21.73 & 0.79 & 50.4M & $\approx$31'' \\
			\hline
			ResNet50 & 64 & 56.78 & 5.25 & 25.6M & $\approx$22'' \\
			SIRe-ResNet50 & 64 & 48.98 & 4.60 & 59.4M & $\approx$38'' \\
			\hline
			GoogleNet & 64 & 43.32 & 1.31 & 13.0M & $\approx$14'' \\
			SIRe-GoogleNet & 64  & 39.51 & 1.28 & 17.0M & $\approx$18'' \\
			\hline
		\end{tabular}
	}
	\caption{Comparison between several literature networks and their SIRe-extended version on the F-MNIST dataset.}
	\label{tab:sire_networks_fmnist}
\end{table}

\begin{table}[t]
	\centering
	\resizebox{\columnwidth}{!}{%
		\begin{tabular}{|l|c|c|c|r|c|}
			\hline
			Model & \thead{Batch\\Size} & \thead{Top-1\\Error\%} & \thead{Top-5\\Error\%} & Params & \thead{Training\\Time} \\
			\hline
			\hline
			Baseline-CNN & 64 & 7.05 & 0.28 & 7.9M & $\approx$121' \\
			SIRe-CNN & 64 & 6.17 & 0.25 & 14.0M & $\approx$207' \\
			\hline
			VGG16 & 64 & 6.08 & 0.29 & 138.3M & $\approx$190' \\
			SIRe-VGG16 & 64 & 4.94 & 0.28 & 160.9M & $\approx$342' \\
			\hline
			VGG19 & 64 & 6.02 & 0.30 & 143.6M & $\approx$201' \\
			SIRe-VGG19 & 64 & 4.95 & 0.31 & 176.8M & $\approx$393' \\
			\hline
			ResNet34 & 64 & 4.72 & 0.10 & 21.8M & $\approx$437' \\
			SIRe-ResNet34 & 64 & 3.99 & 0.10 & 50.4M & $\approx$741' \\
			\hline
			ResNet50 & 64 & 4.57 & 0.11 & 25.6M & $\approx$465' \\
			SIRe-ResNet50 & 64 & 3.82 & 0.10 & 59.4M & $\approx$798' \\
			\hline
			GoogleNet & 64 & 5.16 & 0.24 & 13.0M & $\approx$324' \\
			SIRe-GoogleNet & 64 & 4.09 & 0.18 & 17.0M & $\approx$392' \\
			\hline
		\end{tabular}
	}
	\caption{Comparison between several literature networks and their SIRe-extended version on the CIFAR10 dataset.}
	\label{tab:sire_networks_cifar10}
\end{table}

\begin{table}[t]
	\centering
	\resizebox{\columnwidth}{!}{%
		\begin{tabular}{|l|c|c|c|r|c|}
			\hline
			Model & \thead{Batch\\Size} & 
			\thead{Top-1\\Error\%} & \thead{Top-5\\Error\%} & Params & \thead{Training\\Time} \\
			\hline
			\hline
			Baseline-CNN & 32 & 43.12 & 24.53 & 7.9M & $\approx$16h \\
			SIRe-CNN & 16 & 34.51 & 22.94 & 14.0M & $\approx$27h \\
			\hline
			VGG16 & 32 & 38.13 & 20.67 & 138.3M & $\approx$20h \\
			SIRe-VGG16 & 16 & 30.65 & 17.20 & 160.9M & $\approx$107h \\
			\hline
			VGG19 & 16 & 38.07 & 20.68 & 143.6M & $\approx$31h \\
			SIRe-VGG19 & 8 & 30.96 & 17.19 & 176.8M & $\approx$155h \\
			\hline
			ResNet34 & 16 & 35.65 & 19.98 & 21.8M & $\approx$84h \\
			SIRe-ResNet34 & 8 & 29.80 & 19.01 & 50.4M & $\approx$149h \\
			\hline
			ResNet50* & 8 & 35.04 & 19.95 & 25.6M & $\approx$88h \\
			SIRe-ResNet50* & 4 & 29.53 & 18.97 & 59.4M & $\approx$164h \\
			\hline
			GoogleNet* & 8 & 37.77 & 22.18 & 13.0M & $\approx$66h \\
			SIRe-GoogleNet* & 4 & 31.82 & 20.69 & 17.0M & $\approx$97h \\
			\hline
		\end{tabular}
	}
	\caption{Comparison between several literature networks and their SIRe-extended version on the Caltech256 dataset. Marked (*) models were trained and tested on an Nvidia RTX 3080 with 10 GB of RAM.}
	\label{tab:sire_networks_caltech256}
\end{table}

\subsection{SIRe-Networks Performance Evaluation}
\label{sub:sire_networks}
To complete the SIRe methodology evaluation, several networks were extended to implement the interlaced multi-task learning approach. Notice that while the proposed method can be theoretically applied to any given model performing classification, we selected a few well-known and good performing architectures to both validate the methodology as well as to provide more insights on its strengths and weaknesses.
In particular, the chosen models are the VGG \cite{simonyan2014very}, from which our base CNN takes inspiration, with its VGG16 and VGG19 variants employing the batch normalization; ResNet \cite{he2016deep}, that already implements short residual connections, with its ResNet34 and ResNet50 implementations; and GoogleNet \cite{szegedy2015going}, which reiterates the classification task in its internal layers to mitigate the vanishing gradient and reinforce the final output. 

To correctly apply the SIRe methodology to the aforementioned networks, the various SIRe components need to be integrated with the underlying architectures that might, on the other hand, have articulated solutions (e.g., inception layers). 
Notice that analogously to the presented CNN, in this work we invert all operations in a given $AE$ path of an extended network to recreate the input image, i.e., we pair each max pooling layer with an upsampling one, and convolutions with their transposed implementation; short residual and long skip connections are subsequently implemented to increase the information propagation.
In particular, concerning VGG models, the SIRe extension follows the same approach proposed for the CNN in Fig.~\ref{fig:sire_network}, inverting each path in conjunction with the max pooling operations. Similarly, short residual connections and long skip connections are subsequently defined to obtain the full SIRe-VGG architecture. Notice that VGG models also implement a fifth max pool operation to create a single 512-D feature vector used for classification.
%\color{red}
Therefore, there will be five distinct $AE$s inverting the various blocks in the original architecture
\color{black}
for these extended models (i.e., SIRe-VGG16 and SIRe-VGG19). 
The extended SIRe-VGG16 model is illustrated in Fig.~\ref{fig:sire_vgg16}.
In relation to the ResNet architectures, the base models already implement residual connections through residual blocks. Consequently, to obtain their SIRe version, the backbone network is extended only via the interlaced multi-task learning and long skip connections. Notice that residual blocks are employed also in the various $AE$s but with transposed convolutions to fully invert the various operations. Moreover, similarly to VGG models, the ResNet architecture also implements 5 down sampling operations, 
%\color{red}
resulting in five distinct $AE$s
\color{black}
for both extended models (i.e., SIRe-ResNet34 and SIRe-ResNet50). 
The extended SIRe-ResNet34 model is illustrated in Fig.~\ref{fig:sire_resnet34}.
Lastly, regarding the GoogleNet model, $AE$s were once again attached to each of the 4 max pool operations. Furthermore, inception layers were only partially inverted due to their internal structure. Specifically, all convolutions were substituted via their transposed, while the max pooling operation inside each $AE$ inception layer was left as is for two reasons:
first, the operation is only partially invertible, since it loses information about non-maxima values; and second, in the original inception implementation the filter size remains fixed, therefore allowing for the same max pool operation to be employed in the inverted inception layer. Notice, moreover, that the auxiliary tasks were left untouched, thus resulting in 4 interlaced $AE$s, used for the input image structure preservation, and 2 original internal classifications for a total of 6 internal tasks in the extended model (i.e., SIRe-GoogleNet).

Concerning the evaluation, we compare several base models and their SIRe extension on five datasets. More specifically, we start from CIFAR-100 to assess the proposed method effectiveness on a dataset containing a medium number of classes. We then report experiments on MNIST, Fashion-MNIST, and CIFAR-10 to analyze our method on less complex datasets with a smaller number of classes. Finally, we conclude this evaluation using Caltech-256, which can fully highlight both strengths and weaknesses of the proposed approach. All results are reported in Tables~\ref{tab:sire_networks_cifar100}, \ref{tab:sire_networks_mnist}, \ref{tab:sire_networks_fmnist}, \ref{tab:sire_networks_cifar10}, and \ref{tab:sire_networks_caltech256}. Notice that for each dataset, the corresponding test protocol described in Section~\ref{sub:implementation_details} is employed to assess the various models to provide a fair comparison. 
Starting from CIFAR-100, we report the results obtained on this dataset in Table~\ref{tab:sire_networks_cifar100}.
As shown, the presented methodology improves all architectures for both top-1 and top-5 error metrics; with the former obtaining sensibly higher performance gains (i.e., up to 9\% and 3\% for top-1 and top-5 metrics, respectively). Similarly to the ablation studies, this discrepancy is easily explained by samples moving from a top-5 selection to a top-1 due to the input image information preservation from the $AE$s, i.e., a given object was already close to being correctly classified and the extra information allowed for the right decision. 
What is more, while not reported due to negligible results differences, a good performing $\lambda$-value is close to 0.2 also for other SIRe extended networks;  further confirming both the findings and the discussion presented in Section~\ref{sub:ablation_study}.
On a different note, while SIRe-extended networks can perform better with respect to their original implementation, there is a significant trade-off between the improved performance and the number of parameters required to implement the SIRe methodology. 
Notice, however, that all extra parameters are employed to exclusively address the input image reconstruction task, therefore highlighting the proposed extension effectiveness in improving the original task. Furthermore, the extra parameters amount is highly dependent on the underlying architecture. For instance, SIRe-ResNet50 requires a $\approx$132\% parameter number increase, while SIRe-VGG19 only a $\approx$23\% increment to implement the SIRe methodology. Observe that this lower amount, in correlation with the total parameter quantity shown in Table~\ref{tab:sire_networks_cifar100}, is a consequence of most parameters being associated to the classifier in the original VGG19 architecture.
We also note that the internal structure of a given network directly affects the total training time. For instance, even though the VGG19 has roughly 6x the number of parameters of a ResNet50 (i.e., 143M against 25M), it still requires less time to be fully trained according to the presented test protocol. 
Moreover, the trainable number of parameters also influences the time required to train a given network and, consequently, its SIRe extension (e.g., up to $\approx$70\% for the selected models). 

Similar results can be observed also on datasets with a lower number of classes, i.e., MNIST (Table \ref{tab:sire_networks_mnist}), Fashion-MNIST (Table \ref{tab:sire_networks_fmnist}), and CIFAR-10 (Table \ref{tab:sire_networks_cifar10}), where the SIRe-extended models outperform the corresponding baseline architectures on each dataset. An interesting aspect of the first two of these collections is that the SIRe-extended models achieve considerably lower Top-1 error rates even though they are trained using constrained settings, i.e., 1280 samples for five epochs. The improved performance indicates that the information preserved from the input by the $AE$s can affect the classification task and enables the model to reach convergence faster, even on datasets with fewer classes. This result is particularly noticeable on more complex architectures, i.e., ResNet50 and VGG19, that have higher error rates as they require more epochs to reach convergence with the proposed constraints, due to their design. Unlike the first two datasets, on CIFAR-10, where the networks are trained thoroughly, the error rate discrepancy between base models and their SIRe-extended version is reduced as a consequence of the low number of classes. Nevertheless, the SIRe-networks still outperform the corresponding base models independently of the underlying architecture. 
In this context, notice that even though the Top-5 error of the base models is low on all three datasets, the SIRe-extended architectures can achieve better performance also on this metric. Regardless, the increased training time can already be observed even on this constrained scenario, and becomes non-negligible on the CIFAR-10, i.e., Table \ref{tab:sire_networks_cifar10}, where several SIRe networks required $\approx$1.7x the training time compared with their base version.  

Unlike datasets containing a small or a medium number of samples, where the only drawback of the SIRe methodology lies in the increased training times, the experiments on Caltech-256 also expose another weakness. As reported on Table~\ref{tab:sire_networks_caltech256}, it can be observed that the models use the same number of parameters across all datasets. However, due to the input size of Caltech-256 (i.e., 224$\times$224), the SIRe-extended models require a lower batch size in order to be executed, which directly results in a $\approx$5.3x time increase for some architecture, e.g., SIRe-VGG16. Moreover, this extreme degradation of the training time lead to some of the extended models requiring higher performance hardware. As a matter of fact, SIRe-ResNet50 and SIRe-GoogleNet, contrary to their base versions, could not be trained on the configuration presented in 
%\color{red}
Section~\ref{sub:implementation_details}. 
\color{black}
Notice that, to show a fair comparison, the reported performances refer to experiments carried out on the same hardware even for the base models, although the original architectures, albeit with a low batch size, could still be run on the GeForce GTX 1070.
In conclusion, while the proposed SIRe extension can be theoretically applied to any network performing classification, further investigations on possible reduced interlaced multi-task implementations (e.g., less $AE$ layers) are required to simplify, without loss of generality, the SIRe extended network training procedure; an improvement that would enable the presented methodology application to the ever more complex architectures being developed, but which is left as future work.

\section{Conclusion}
In this paper we introduced the SIRe methodology, that allowed us to improve the classification capabilities of a given network by preserving information from the input image through an interlaced multi-task learning strategy. Furthermore, the latter was refined via long skip and short residual connections to increase the quality of the reconstructed images.
Moreover, both strengths and weaknesses for the proposed approach, i.e., a significant classification performance boost and a required training time increase, were presented and discussed through several ablation studies. Finally, the SIRe methodology was also applied to different well-known literature works, validating the proposed strategy by means of improved performances on all of the extended architectures when tested on five different datasets containing a low, a medium, and a high number of classes.

As future work we plan to investigate possible strategies to reduce the amount of parameters required to implement the SIRe approach in even more complex architectures, without any loss of generality for the input image reconstruction and information preservation rationale. Furthermore, other extensive experiments will be carried out to explore the effects of different and multiple interlaced tasks with respect to the classification one, to ultimately obtain ever improving feature abstractions.
In this context, a thorough analysis of the loss landscape will also be performed by following the ideas presented in \cite{li2017visualizing} to have an in-depth understanding of how the SIRe components and the multiple interlaced tasks affect the gradient evolution at training time.

\section*{Acknowledgments}
This work was supported in part by the MIUR under grant “Departments of Excellence 2018–2022” of the Department of Computer Science of Sapienza University.

\bibliographystyle{elsarticle-num} 
\bibliography{refs}


	
	
	
	% \end{thebibliography}
\end{document}